\documentclass[runningheads]{llncs}
\usepackage{graphicx}
\usepackage{booktabs}
\usepackage{textgreek}
\usepackage{float}
\usepackage{hyperref}

\hypersetup{colorlinks=true,
colorlinks=true,
     linkcolor=black,
     filecolor=black,
     citecolor = black,      
     urlcolor=blue,
     }

\begin{document}
%
\title{Age-Related Differences in the Perception \\ of Eye-Gaze from a Social Robot%
\thanks{\textbf{This is the author's version of a paper accepted for publication in the Proceedings of the 13th International Conference on Social Robotics (ICSR 2021). The final authenticated publication is available online at: \url{https://doi.org/10.1007/978-3-030-90525-5_30}}}}

\titlerunning{Age-Related Diff. in the Perception \\ of Eye-Gaze from a Social Robot}

%
\author{Lucas Morillo-Mendez\orcidID{0000-0001-7339-8118} \and
Martien G.S. Schrooten\orcidID{0000-0002-9462-0256} \and
Oscar Martinez Mozos\orcidID{0000-0002-3908-4921}}
\authorrunning{Morillo-Mendez et al.}
%
\institute{Örebro University, Fakultetsgatan 1, 702 81, Örebro, Sweden\\
\email{\{lucas.morillo, martien.schrooten, oscar.mozos\}@oru.se}}
\maketitle              
%

\begin{abstract}

There is an increasing interest in social robots assisting older adults during daily life tasks. In this context, non-verbal cues such as deictic gaze are important in natural communication in human-robot interaction. However, the sensibility to deictic-gaze declines naturally with age and results in a reduction in social perception. Therefore, this work explores the benefits of deictic gaze from social robots assisting older adults during daily life tasks, and how age-related differences may influence their social perception in contrast to younger populations. This may help on the design of adaptive age-related non-verbal cues in the Human-Robot Interaction context.

\keywords{Human-Robot Interaction, Older Adults, Non-Verbal Cues}


\end{abstract}
\section{Introduction}
In the last years, there has been an increasing interest in the use of social robots to assist older adults (OA) during  daily life tasks~\cite{pu2019}. An important cue in the interaction with social robots is non-verbal communication such as deictic gaze~\cite{Mutlu2009,Admoni2016}. Humans use deictic gaze to guide the attention of another person towards a point in the space by looking at it. This communicative signal is key to initiate a shared focus between individuals and to increase the efficiency in collaborative tasks~\cite{Cohen1995}. In addition, deictic gaze is important in OA because it can help to inform age-related differences in human-robot interaction (HRI)~\cite{cani2019}. This is due to the fact that the sensibility to deictic gaze declines naturally with age, reflecting a reduction in social perception in OA \cite{Slessor2016}. For this reason, it is important to explore how deictic gaze is attended in normal ageing when performed by a robot. 

At the same time, it has been reported a need for more studies regarding non-verbal cues in which OA are direct research participants, in contrast to studies where OA act only as beneficiaries, and to further compare the outputs with younger controls~\cite{Zafrani2019}. Therefore, it is interesting to explore the benefits of non-verbal cues from social robots towards OA during collaborative daily life tasks, and how age-related differences may influence their perception of a social robot in contrast to younger populations. Finally, these studies may help improving the design of non-verbal cues in HRI that adapt to age changes.

This work seeks to explore potential age-related differences in the perception of deictic gaze from a social robot when collaborating in tasks inspired by daily life activities. We used a Pepper robot\footnote{\url{https://www.softbankrobotics.com}} given its wide use in research related to HRI. Pepper does not have degrees of freedom in the eyes to reflect human-like gaze, therefore, and in line with previous research~\cite{Admoni2016,Mwangi2018}, we used its head movement to point to objects as a way to reflect deictic gaze.

\begin{figure}
\centering
\includegraphics[width= 0.55\linewidth]{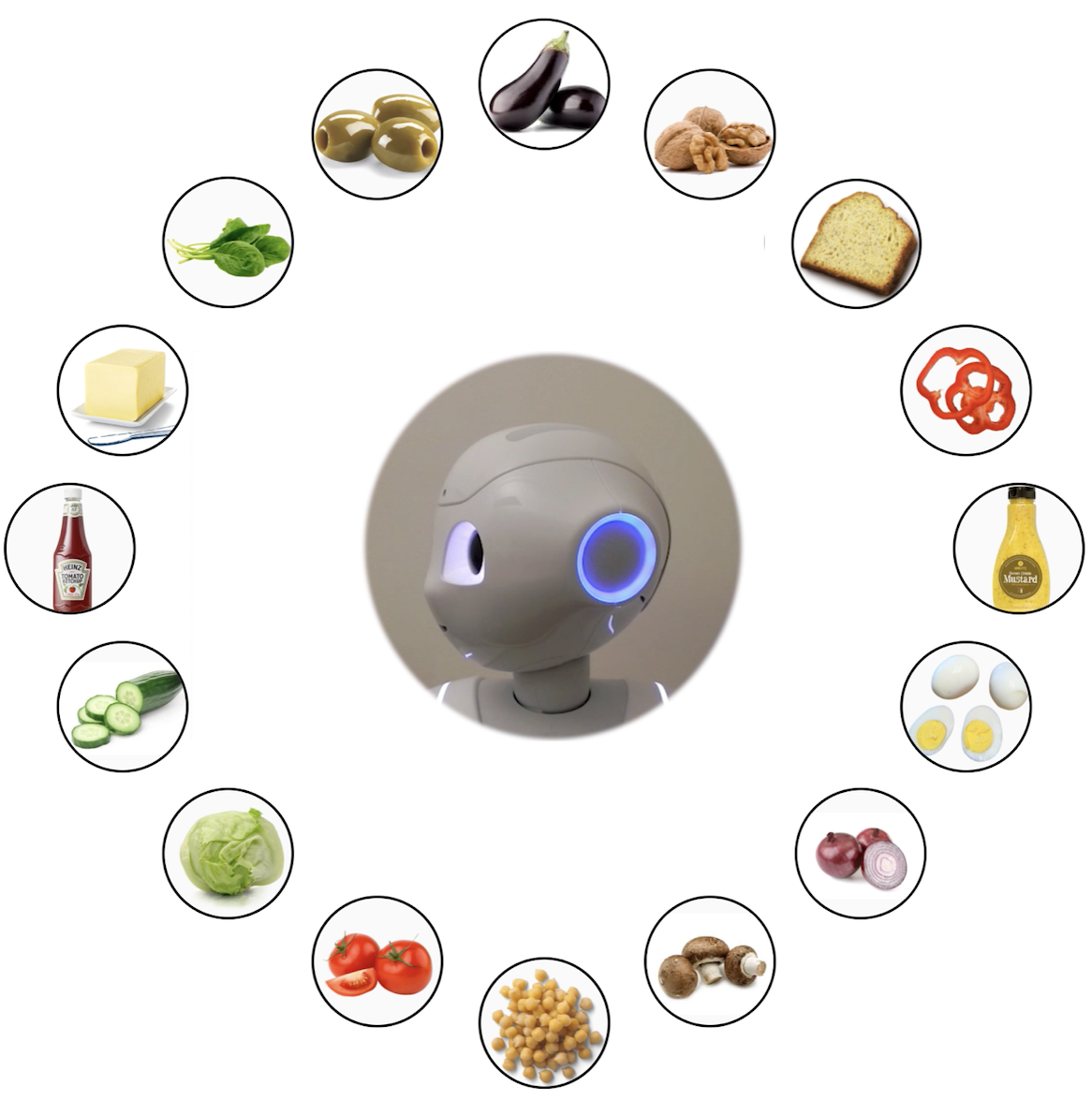}

\caption{Experimental layout.}

\label{fig:pep}
\end{figure}

Due to the Covid-19 global pandemic, we decided to look for safer ways to move research in HRI for OA forward. Therefore, we designed an online controlled collaboration task mimicking an everyday life situation that permitted remote participation. This also allowed us to recruit more than 300 participants, including OA, a high number compared to in-lab experiments. In our experiment, a video of Pepper verbally guided the participants during a task that mimicked the preparation of a sandwich as shown in Fig.~\ref{fig:pep}. In order to compare the potential benefit in the perception of non-verbal cues, the robot switched its behaviour between static-based and deictic gaze-based indications. Then we measured different times during the collaborative tasks to analyze potential differences between age populations. 
An example video with the task is available\footnote{\url{https://youtu.be/6zSgm8jEnCM}}.

Our results show a significant facilitation effect of deictic gaze from a Pepper robot in all the participants independently of their age. These findings show that head movement representing deictic gaze is effective in terms of task performance in general. However, this facilitation effect was not significantly different between the age groups. Moreover, OA's social perception of the robot was less influenced by its deictic gaze behavior. 

\section{Related Work}
Gaze behavior from a robot can affect HRI in different ways. Appropriate gaze from a speaking robot towards a human can positively affect the recall of what have been said \cite{Mutlu2006}, and can regulate the role of the participants in conversations~\cite{Mutlu2009}. Similarly, a robot moving the head away can effectively reflect gaze aversion and can be perceived as more thoughtful by the users~\cite{Andrist2014}. In collaborative scenarios where a human follows guidance from a robot, deictic gaze from the robot can assist the human partner by signalling at objects in space~\cite{Admoni2017}, although the specific benefits differ among studies \cite{Admoni2016,Mwangi2018,Kontogiorgos2019}.


The work in~\cite{Kontogiorgos2019} investigated deictic gaze in a situated human-agent collaboration. The results showed that this non-verbal cue led to higher interaction times and, thus, inferior task performance. In contrast, the work in~\cite{Mwangi2018} found that deictic gaze in the form of head movement from a robot did not affect task completion times, although it helped to reduce the number of errors. This is in line with \cite{Admoni2016}, which suggested that deictic gaze from a robot is not that useful in simple tasks when compared to difficult ones.



Previous research also indicated that eye-gaze following deteriorates with age~\cite{Slessor2016}. Nevertheless, to the best of our knowledge, the influence that gaze from a robot may have on OA has not been explored. In our study we used a similar task as in~\cite{Kontogiorgos2019} since we think it reflects better an interaction that OA can have with an assistive robot. The remote nature of our study limited the interaction between the users and the robot, but allowed us to control the influence of extraneous variables on the main outcome variables, such as the social presence caused by the robot looking at the user, which may lead the users to start a conversation with the robot and get higher completion times as reported in~\cite{Kontogiorgos2019}.

\section{Methods}
\label{sec:methods}

\subsection{Scenario}
\label{sec:scenario}

\begin{figure*}[t]
      \centering
      \includegraphics[width = 1\linewidth]{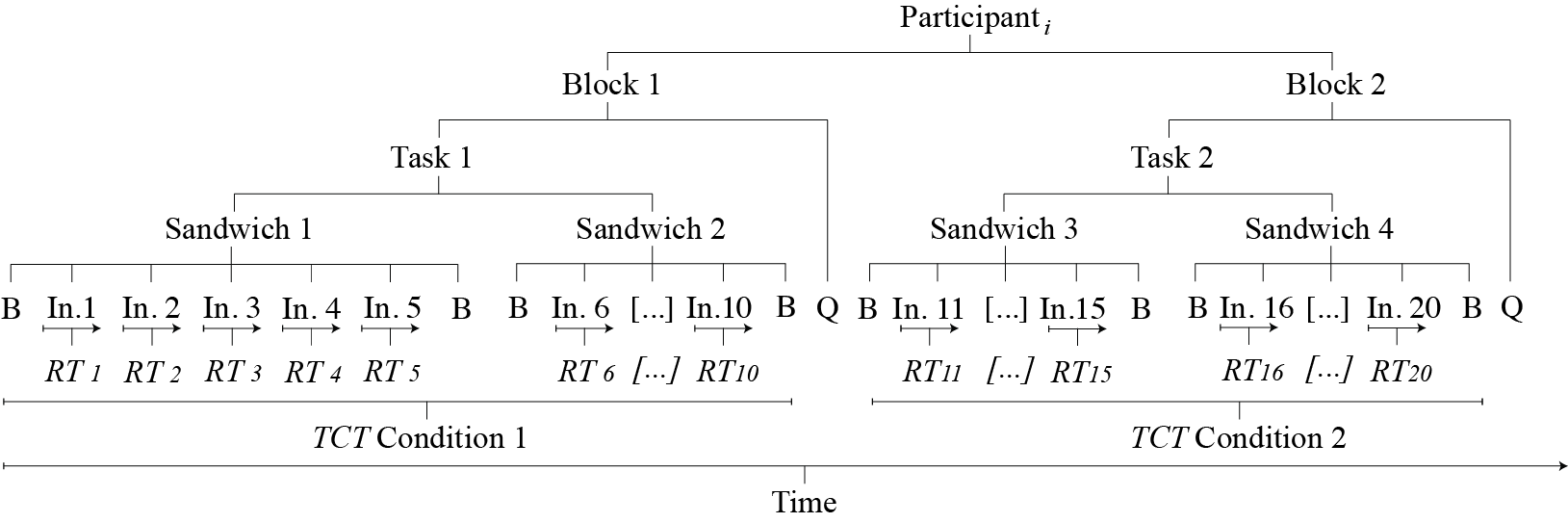}
      \caption{Structure of the experiment for a participant \textit{i}. The letter `B' represents the bread, while `In. x' represents ingredient in position x. `Q' refers to the questionnaire.}
      \label{struct}
\vspace{-0.3cm}
  \end{figure*}

Our experimental scenario featured a video of a pepper robot who verbally guided participants through two everyday-like visual search tasks which consisted on clicking on several ingredients for preparing a sandwich. An example layout of the ingredients and the robot is shown in Fig.\ref{fig:pep}. We defined two robot conditions: a \textit{static robot} (SR) which always looked at the camera while giving instructions, and a \textit{moving robot} (MR) which also featured deictic gaze by moving the head towards the correct ingredient. Each of these conditions defined a task: a SR task, and a MR task. Inside each task, the user had to prepare two sandwiches by following the verbal instructions of the Pepper robot, which named each ingredient and waited for the user to click on it. Therefore each participant prepared two consecutive sandwiches with a a static robot (SR), and two other consecutive sandwiches with a moving robot (MR). The full structure of the experiment is shown in Fig.~\ref{struct}.


A trial consisted on the selection of one ingredient. To obtain a larger amount of trials, the user prepared two sandwiches in each task. The order of the sandwiches within a task was fixed and the order of the ingredients in each sandwich was also fixed to maintain its coherence. The tasks were counterbalanced and started randomly either with SR or MR.

Inside a trial, we defined the reaction time (RT) as the time span between the moment the robot started naming one ingredient and the moment the participant clicked on that ingredient. The RT of the \textit{bread} was excluded because of its high predictability (always first/last ingredient). A trial was correct when the mentioned ingredient was clicked. The final number of trials per task in which RT was used was ten: five ingredients for each sandwich. In addition, we defined the task completion time (TCT) as the time needed by a participant to finish one task (two sandwiches). 



\subsection{Materials}
\label{sec:materials}

After each task, participants were presented with a set of questionnaires and subscales. First, the mental demand subscale of the NASA-Task Load Index (NASA-TLX) \cite{Hart1988} was used for assessing the mental demand between robot conditions. Second, the Godspeed Questionnaire Series \cite{Bartneck2009} was used to measure the perceived anthropomorphism of the robot by the user. Here, we used a modified version of the anthropomorphism questionnaire due to an irrelevant item for the context of this study (\textit{moving rigidly-elegantly}). Moreover, we added the item \textit{mechanical-organic} as in \cite{Carpinella2017}. Finally, the RoSAS \cite{Carpinella2017} scale was used to measure the perception of warmth, competence, and discomfort caused by a robot. We added two extra questions at the end: Q1) \textit{Did you notice any difference between the robots in the tasks?} to check whether the person was aware of the difference between robot conditions; and Q2) \textit{Which robot you prefer from the ones you interacted with?} to check their preferred condition (SR or MR). 


This study was built using \textit{Labvanced} \cite{Finger2017}, an online tool for designing and remotely distributing experiments on human cognition. 
The language was Spanish since all participants were fluent in that language. 


\subsection{Experimental Design}
\label{sec:exp}

Our study followed a 2x2 mixed design with two robot conditions, SR and MR (within-subject), and two age groups, adults (A) and OA (between-subjects). The age range in the A was $>18$ and $<65$ years, and in OA it was $>64$. This division was based on the working retirement age.

The presentation order of the blocks (see Fig.~\ref{struct}) was counterbalanced and participants were randomly assigned to one of the two possible orders. To explore a possible differential effect of deictic gaze from a robot between our groups of age in times and social attributes towards the robot, we focused on the analysis of the interactions between age and robot condition. 




\begin{table}[t]
\centering
\caption{Sample description} 
\label{tab:table}
\resizebox{\columnwidth}{!}{
\begin{tabular}{@{\extracolsep{4pt}}lccccccccccc}
\toprule   
{} & \multicolumn{2}{c}{Age (years)} & \multicolumn{3}{c}{Gender} & \multicolumn{3}{c}{Comfort w/ computers} & \multicolumn{3}{c}{Had seen Pepper before} \\
 \cmidrule{2-3} 
 \cmidrule{4-6} 
 \cmidrule{7-9} 
 \cmidrule{10-12} 
 Group & Mean & SD & M & F & Other & No & Not Sure & Yes & No & Not Sure & Yes \\ 
\midrule
OA & 69.3 & 3.8 & 76 & 74 & 0 & 1 & 7 & 142 & 108 & 26 & 16  \\
A & 53.4 & 12.1 & 45 & 80 & 1 & 1 & 11 & 113 & 84 & 24 & 18  \\
\bottomrule
\end{tabular}%
}
\end{table}


\subsection{Sample}
\label{sec:sample}

We performed a G*Power analysis \cite{Faul2007} to calculate a minimum sample size that allowed a expected power (1 - \textbeta) of 0.80 to detect a small effect size of $f=0.25$ ($\eta ^2 = 0.06$) between age groups. A total of 329 participants took part in the experiment, of which 53 were excluded due to incomplete data. A summary of the sample is shown in Table \ref{tab:table}. Potential participants were contacted via mailing lists from Spanish universities with adult education programs. The conditions to participate were to be fluent in Spanish, have normal or corrected-to-normal vision, use headphones in order to reduce potential external noise, and to be cognitively healthy. Participants gave written informed consent in accordance to the Declaration of Helsinki and were informed about research goals. Participation was voluntary and no personal data that allowed their identification were obtained. In addition, approval was obtained from the corresponding program coordinators at each university. A chi-squared test showed no significant differences between the age groups in their level of education, their previous knowledge of Pepper, and their experience with computers. The proportion of men in the A group was significantly lower ($\chi^2(1)=5.9, p=0.014$) than in OA.

\subsection{Procedure}
\label{sec:procedure}

The experiment started by asking participants to calibrate the volume of their headphones. Then, they were informed in the consent form about the study and the possibility of ending it at any time and filled a sample information questionnaire. Before the real experiment started, participants had time to get familiar with the interface. To reduce external influences, participants were encouraged to avoid distractions and to be rested before starting. To favour this, we kept the experiment short and they were informed about its duration, fifteen minutes. For the main tasks, they were also encouraged to perform as well as they could, but without explicit instructions about being fast. This was to maintain the everyday nature of the task in contrast to a classic computerized experiment. In addition, they were not warned about the difference between the robot conditions (SR or MR). 

\section{Results}
\label{sec:results}

We first present reaction times (RT) and task completion times (TCT) between the robot conditions (SR, MR), age groups (OA, A), and the combination of both. Potential noise in the time measures from participants due to external factors such as computer, browser, or operative system, was corrected using metadata provided by Labvanced. To analyze the RT data, we used the median RT of the correct trials within each task per participant (Fig. \ref{struct}). The percentage of incorrect, and thus excluded, trials was 2.24\%. Due to violations of assumptions for the mixed ANOVA test, we analyzed the data using a Mixed Robust ANOVA test with 20\% trimmed means and 2000 bootstrapped samples. 

\begin{figure}[t]
      \centering
      \includegraphics[width = 1\linewidth]{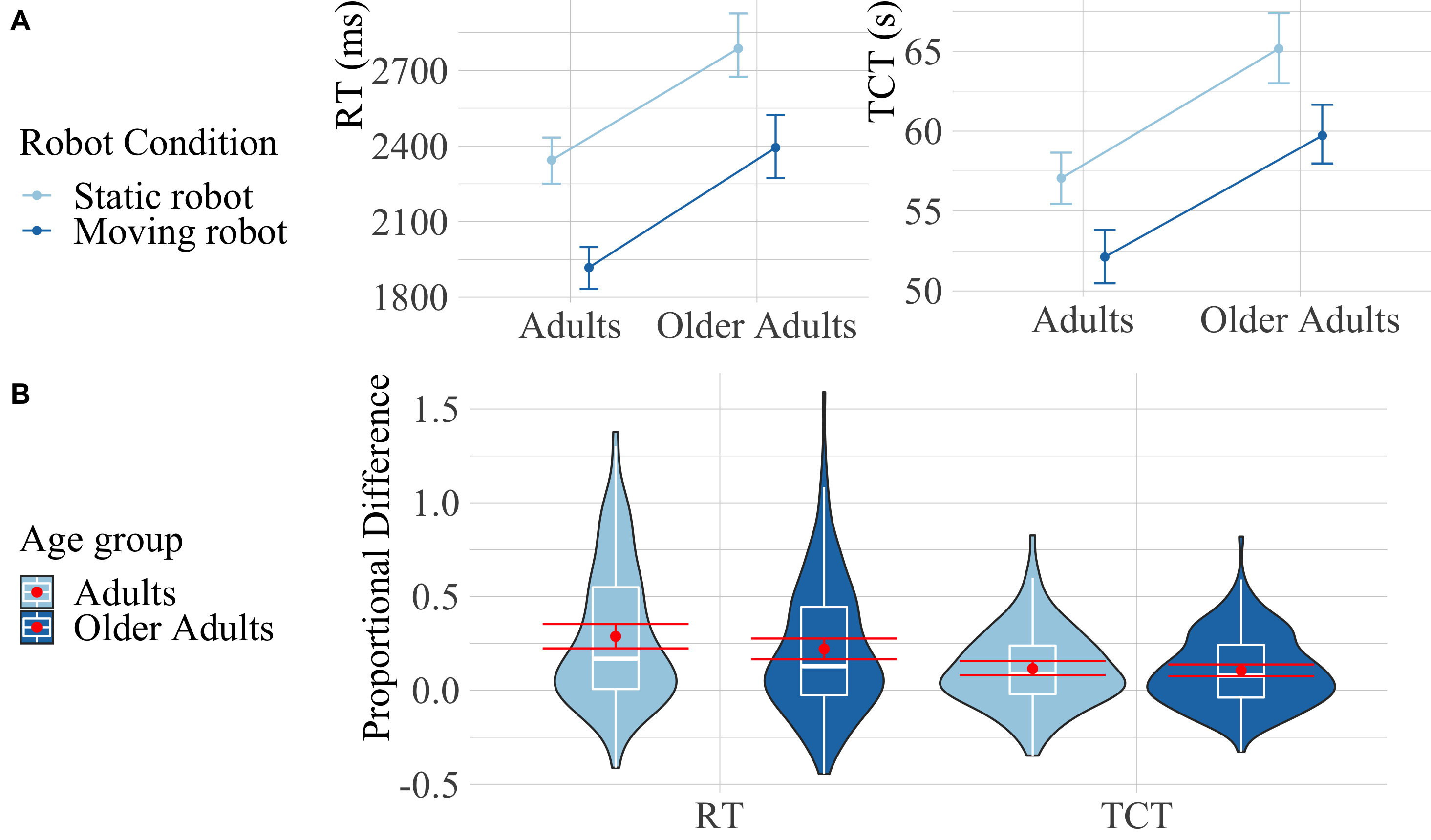}
       \caption{A) Mean Reaction Time (left) and Task Completion Time (right) for each group. B) Violin plots with means in red of the proportional differences between robots. Error bars show 95\% bootstrapped confidence intervals.}
      \label{resul}
      \vspace{-0.5cm}
\end{figure}

The RT for different age groups and conditions are shown in Fig.~\ref{resul}A(left). The results are reported in Table~\ref{tab:table2}. Values were significantly different between the robot conditions, showing a facilitation effect for the MR condition. We also found a main effect of age that showed higher RT for OA. We did not find any interaction effects in RT between robot condition and age group. However, interaction analyses in which RT are involved are not easily interpretable in this case. It has been shown that overall RT slowing, as in OA, increases RT's differences inside this group~\cite{Chapman1994}. Following~\cite{Slessor2016}, the strength of the facilitation effect was calculated as a proportional difference score $(RT_{SR} - RT_{MR}) / RT_{MR}$. An independent robust t-test (trim = 0.2, samples = 2000) did not show significant differences in the strength of the facilitation effect caused by the MR in RT between age groups (Fig.~\ref{resul}B and Table~\ref{tab:table2} ), i.e., the magnitude of the help provided by the MR in terms of how fast the ingredients were clicked was similar between the age groups. 
  
The TCT for different age groups and conditions are shown in Fig.~\ref{resul}A(right). Means, standard deviations and p-values for TCT are also reported in Table~\ref{tab:table2}. Results show a facilitation effect for the MR condition in TCT. Furthermore, TCT was also higher for OA. We could not find any interaction effects between robot condition and age group. The strength of the facilitation effect was also calculated as a proportional difference score $(TCT_{SR}-TCT_{MR})/TCT_{MR}$. A robust t-test did not show significant differences in the strength of the facilitation effect caused by the MR in TCT between age groups (Fig. \ref{resul}B and Table \ref{tab:table2}). Finally, the magnitude of the help provided by the MR in terms of how fast the task was performed was similar between the age groups. 



We now present the scores of the different questionnaires and scales from Sect. \ref{sec:materials}. We used a Robust Mixed ANOVA for all the scores except for anthropomorphism, as it met the assumptions for a regular Mixed ANOVA. Table~\ref{tab:table3} shows the social perception scores and the Cronbachs's $\alpha$ of each construct with the corresponding analyses. We found a significant effect of age by which OA perceived the robots as more anthropomorphic. In addition, the MR scored significantly higher than SR in all the social perception scores except in the discomfort score. This indicates a more positive perception of the MR. Finally, a significant interaction effect showed that anthropomorphism and discomfort remained more invariant in the OA group as a result of the deictic behavior of the robot. 

\begin{table}[t]
\centering
\caption{Main and interaction effects on the times} 
\label{tab:table2}
\resizebox{\columnwidth}{!}{
\begin{tabular}{@{\extracolsep{19pt}}lllllll}
\toprule   
{} & RT (ms) & TCT (s) & \% Facilitation (RT) & \% Facilitation (TCT) \\
\midrule
Age & $p<$0.001 (***) & $p<$0.001 (***) &  $p$=0.31 & $p$=0.66 \\
 $M^{A}$ & $2131\pm575$ & $53.5\pm9.8$ & $0.29\pm0.36$ & $0.12\pm0.2$ \\
 $M^{OA}$& $2590\pm808$ & $61\pm12.5$ & $0.22\pm0.35$ & $0.11\pm0.2$ \\
\midrule
Robot & $p<$0.001 (***) & $p<$0.001 (***) & - & - \\
 $M^{SR}$ & $2585\pm721$ & $61.5\pm12.3$ &  &  \\
 $M^{MR}$ & $2176\pm716$ & $56.3\pm11.1$ &  &  \\
\midrule
Age*Robot & $p$=0.2 & $p$=0.7& - & - \\
 $M^{SR-MR}_{A}$ & $426\pm582$ & $4.9\pm10.7$ &  &  \\
 $M^{SR-MR}_{OA}$ & $393\pm744$ & $5.4\pm11$ &  &  \\
\bottomrule
\end{tabular}%
}
\bigskip
\centering
\caption{Main and interaction effects on the social perception scores} 
\label{tab:table3}
\resizebox{\columnwidth}{!}{
\begin{tabular}{@{\extracolsep{19pt}}lllllll}
\toprule   
{} & Anth. ($\alpha$ = .88). & Warmth ($\alpha$ = .87) & Compt. ($\alpha$ = .84) & Discom. ($\alpha$ = .76) \\
\midrule
Age & $p$=0.02 (*) & $p$=0.051 &  $p$=0.26 & $p$=0.88 \\
 $M^{A}$ & $2.66\pm0.94$ & $2.47\pm0.8$ & $3.64\pm0.67$ & $1.71\pm0.58$ \\
 $M^{OA}$& $2.89\pm0.77$ & $2.56\pm0.8$ & $3.65\pm0.73$ & $1.73\pm0.58$ \\
\midrule
Robot & $p<$0.001 (***) & $p$=0.003 (**) & $p<$0.001 (***) & $p<$0.001 (***) \\
 $M^{SR}$ & $2.7\pm0.92$ & $2.42\pm0.8$ & $3.49\pm0.76$ & $1.76\pm0.6$ \\
 $M^{MR}$ & $2.87\pm0.7$ & $2.62\pm0.79$ & $3.8\pm0.61$ & $1.68\pm0.55$ \\
\midrule
Age*Robot & $p$=0.03 (*) & $p$=0.85 & $p$=0.47 & $p$=0.003 (**) \\
 $M^{SR-MR}_{A}$ & $-0.27\pm0.77$ & $-0.25\pm0.67$ & $-0.39\pm0.71$ & $0.173\pm0.58$ \\
 $M^{SR-MR}_{OA}$ & $-0.08\pm0.66$ & $-0.15\pm0.62$ & $-0.25\pm0.64$ & $0\pm0.5$ \\
\bottomrule
\end{tabular}%
}
\end{table} 


We also analysed the subset of all participants who did not notice the difference between the robot conditions by answering \emph{No} to Q1, (see Sect.\ref{sec:materials}). This was done to test whether the deictic gaze of the robot affected the reaction times even if the participants were unaware of that movement. In this subset (a total of 116), we found a main effect of age group on RT at $p<0.001$, but we did not find a robot gaze nor interaction effects in this case. We also found a main effect of age group on TCT at $p<0.001$. However, we did not find a robot condition effect nor interaction effect. No effects were found in any of the subjective scores. Inside that subset, a Chi-squared test showed significant differences between age groups in the percentages of participants, A=30\%, OA=52\% ($\chi^2(1)=12.5, p < 0.001$). 

We also analysed the participants who expressed a preference for a robot (SR or MR) by answering to \textit{Q2}. From a total of 163 answers, $78.5\%$ chose the MR. Inside the participants choosing the MR, the differences between age groups, A=82\%, OA=77\%, were not significant ($\chi^2(1)=2.47, p = 0.11$).  
Finally, with respect to the mental demand scale, we found no effects of age ($p=0.07$), robot ($p=0.11$), and interaction ($p=0.33$) ($M=4.18\pm3.75$).

\section{Discussion and Conclusion}

This work sought to explore potential age-related differences in the perception of visual cues from a social robot. We focused on the influence of deictic gaze during a collaborative tasks inspired by daily life activities. We found a facilitation effect of deictic gaze from a Pepper robot in all the participants independently of their age. Given this facilitation effect for both time scales, our main interest was to find if its magnitude was different between age groups. 
Our results showed that the facilitation effect from the deictic gaze was not significantly different between age groups, neither in TCT nor RT (Fig.~\ref{resul}; Table \ref{tab:table2}). To further investigate the effect of deictic gaze
future research could add more conditions like a human face or non-social signalling.


OA scored higher in anthropomorphism regardless of the robot condition. 
There was also an increase in all the social perception scores as a result of the robot deictic gaze (Table \ref{tab:table3}). In addition, the MR was chosen as favourite. These results support previous notions that appropriate social behaviors improve the acceptability of social robots~\cite{Feingold-Polak2018}. 
In addition, the participants indicated a low mental demand ($M=4.18$ out of $21$) when performing the tasks, independently of the robot condition. Moreover, we found an interaction effect in the scores of anthropomorphism and discomfort caused by the robot (Table~\ref{tab:table3}) which varied less between robot conditions in OA. This interactions suggests a different perception of deictic gaze from a robot by OA. 

A proportion of $42\%$ of the participants reported not detecting the differences between the robot conditions. Inside this group we only found age effects on RT and TCT (Sec.~\ref{sec:results}). 
Although these participants did not notice the differences between robot conditions, the incorrect trial ratio in our sample remained low ($2.24\%$). 
Probably they did not take advantage of the deictic gaze towards the correct ingredient. 

Inside this $42\%$ the proportion of OA was significantly higher than A. 
The work in~\cite{Slessor2016} shows that OA present a decline in eye-gaze following.  
However, this might as well reflect a broader cognitive decline. The negative answer could also indicate a difficulty in remembering the differences between robot conditions. 

 In conclusion, we found a facilitation effect of deictic gaze from a Pepper robot in all the participants independently of their age. These findings show that head movement representing deictic gaze is effective in terms of task performance in general. However, this facilitation effect was not significantly different between the age groups which means that A do not benefit more than OA. Moreover, we found age-related differences in the social perception scores as result of the deictic gaze from the robot. Nevertheless, OA seem to be less reactive to deictic gaze when it comes to subjective scores of anthropomorphism and discomfort caused by the robot. Future research should add human and/or non-social controls to better inform the differences between the perception of human and robot gaze cues. The result of these studies could be valuable in future designs of non-verbal cues from robots in HRI.


\subsubsection{Acknowledgments.}
This work is funded by the EU Horizon 2020 research and innovation programme under the Marie Skłodowska-Curie grant agreement No 754285, by the Wallenberg AI, Autonomous Systems and Software Program (WASP) funded by the Knut and Alice Wallenberg Foundation, and by the RobWellproject (No RTI2018-095599-A-C22) funded by the Spanish Ministerio de Ciencia, Innovación y Universidades. We want to thank the universities that helped us with the sample recruiting process: Complutense University of Madrid, University Carlos III of Madrid, University of Murcia and University of Alicante. The main author wants to express his gratitude to Neziha Akalin and Estefanía Sánchez-Pastor for their support in the initial stages of this study.

\bibliographystyle{splncs04}
\bibliography{bib}

@inproceedings{Finger2017,
address = {Cologne, Germany},
author = {Finger, Holger and Goeke, Caspar and Diekamp, Dorena and Standvo{\ss}, Kai and K{\"{o}}nig, Peter},
booktitle = {Proc. Int Conf. IC2S2 ({IC2S2}'17)},
title = "LabVanced: A Unified JavaScript Framework for Online Studies",
year = {2017}
}

@article{Feingold-Polak2018,
author = {Feingold-Polak, Ronit and Elishay, Avital and Shahar, Yonat and Stein, Maayan and Edan, Yael and Levy-Tzedek, Shelly},
journal = {Paladyn},
number = {1},
pages = {183-192},
publisher = {De Gruyter},
title = "Differences between young and old users when interacting with a humanoid robot: A qualitative usability study",
volume = {9},
year = {2018}
}

@article{Mwangi2018,
author = {Mwangi, Eunice and Barakova, Emilia I. and D{\'{i}}az-Boladeras, Marta and Mallofr{\'{e}}, Andreu Catal{\`{a}} and Rauterberg, Matthias},
issn = {18754805},
journal = {Int. J. Soc. Robot.},
number = {3},
pages = {343-355},
title = "Directing Attention Through Gaze Hints Improves Task Solving in Human–Humanoid Interaction",
volume = {10},
year = {2018}
}

@article{Admoni2017,
author = {Admoni, Henny and Scassellati, Brian},
issn = {2163-0364},
journal = {Journal of Human-Robot Interaction},
pages = {25–63},
number = {1},
publisher = {Journal of Human-Robot Interaction},
title = "Social Eye Gaze in Human-Robot Interaction: A Review",
volume = {6},
year = {2017}
}

@inproceedings{Carpinella2017,
  author={C. M. {Carpinella} and A. B. {Wyman} and M. A. {Perez} and S. J. {Stroessner}},
  booktitle={Proc. {ACM/IEEE} Int. Conf. HRI ({HRI}'17)}, 
  title="The Robotic Social Attributes Scale ({RoSAS}): Development and Validation", 
  year={2017},
  pages={254-262},
 address = {New York, USA}
}

@inproceedings{Kontogiorgos2019,
author = {Kontogiorgos, Dimosthenis and Pereira, Andre and Andersson, Olle and Koivisto, Marco and Rabal, Elena Gonzalez and Vartiainen, Ville and Gustafson, Joakim},
booktitle = {Proc. {ACM} Int. Conf. IVA ({IVA}'19)},
isbn = {9781450366724},
pages = {133-140},
title = "The effects of anthropomorphism and non-verbal social behaviour in virtual assistants",
year = {2019},
address = {Paris, France}
}

@article{Hart1988,
author = {Hart, Sandra G. and Staveland, Lowell E.},
issn = {01664115},
journal = {Adv. Psychol},

pages = {139-183},
publisher = {North-Holland},
title = "Development of {NASA-TLX} ({Task Load Index}): Results of Empirical and Theoretical Research",
volume = {52},
year = {1988}
}

@article{Bartneck2009,
author = {Bartneck, Christoph and Kuli{\'{c}}, Dana and Croft, Elizabeth and Zoghbi, Susana},
journal = {International Journal of Social Robotics},
number = {1},
pages = {71-81},
title = "Measurement instruments for the anthropomorphism, animacy, likeability, perceived intelligence, and perceived safety of robots",
volume = {1},
year = {2009}
}

@article{Zafrani2019,
   author = {Oded Zafrani and Galit Nimrod},
   issue = {1},
   journal = {Gerontologist},
   pages = {26-36},
   pmid = {30016437},
   publisher = {Gerontological Society of America},
   title = "Towards a Holistic Approach to Studying Human-Robot Interaction in Later Life",
   volume = {59},
   year = {2019},
}

@book{Cohen1995,
   author = {Simon Baron-Cohen},
   publisher = {The MIT Press},
   title = {Mindblindness: An Essay on Autism and Theory of Mind},
   year = {1995},
   address = {Cambridge}
}

@inproceedings{Admoni2016,
   author = {Henny Admoni and Thomas Weng and Bradley Hayes and Brian Scassellati},
   isbn = {9781467383707},
   issn = {21672148},
   booktitle = "Proc. {ACM/IEEE} Int. Conf. HRI ({HRI}'16)",
   pages = {51-58},
   address = "Christchurch, New Zealand",
   title = "Robot nonverbal behavior improves task performance in difficult collaborations",
   year = {2016},
}

@article{Pu2019,
   author = {Lihui Pu and Wendy Moyle and Cindy Jones and Michael Todorovic},
   issn = {0016-9013},
   issue = {1},
   journal = {The Gerontologist},
   pages = {37-51},
   publisher = {Gerontological Society of America},
   title = "The Effectiveness of Social Robots for Older Adults: A Systematic Review and Meta-Analysis of Randomized Controlled Studies",
   volume = {59},
   year = {2019}
}

@article{cani2019,
   author = {Roser Cañigueral and Antonia F.de C. Hamilton},
   issn = {16641078},
   issue = {mar},
   journal = {Frontiers in Psychology},
   publisher = {Frontiers Media S.A.},
   title = "The role of eye gaze during natural social interactions in typical and autistic people",
   volume = {10},
   year = {2019},
}

@inproceedings{Andrist2014,
   author = {Sean Andrist and Xiang Zhi Tan and Michael Gleicher and Bilge Mutlu},
   isbn = {9781450326582},
   issn = {21672148},
   title = {Conversational gaze aversion for humanlike robots},
   booktitle = {Proc. ACM/IEEE Int. Conf. HRI ({HRI}'14)},
   pages = {25-32},
   address = "Bielefeld, Germany",
   year = {2014}
}

@inproceedings{Mutlu2006,
   author = {Bilge Mutlu and Jodi Forlizzi and Jessica Hodgins},
   isbn = {142440200X},
   booktitle = {Proc. {IEEE-RAS} Int. Conf. HUMANOIDS},
   pages = {518-523},
   title = {A storytelling robot: Modeling and evaluation of human-like gaze behavior},
   year = {2006},
   address = {Genoa, Italy}
}

@inproceedings{Mutlu2009,
   author = {Bilge Mutlu and Toshiyuki Shiwa and Takayuki Kanda and Hiroshi Ishiguro and Norihiro Hagita},
   pages = {61},
   booktitle = {Proc. {ACM/IEEE} Int.Conf. HRI (HRI’09)},
   title = {Footing in human-robot conversations},
   year = {2009},
   address = {La Jolla, California, USA}
}

@article{Slessor2016,
   author = {Gillian Slessor and Cristina Venturini and Emily J. Bonny and Pauline M. Insch and Anna Rokaszewicz and Ailbhe N. Finnerty},
   issn = {1079-5014},
   issue = {1},
   journal = {J. Gerontol. Series B},
   pages = {11-22},
   publisher = {Gerontological Society of America},
   title = {Specificity of Age-Related Differences in Eye-Gaze Following: Evidence From Social and Nonsocial Stimuli},
   volume = {71},
   year = {2016},
}

@article{Faul2007,
author = {Faul, Franz and Erdfelder, Edgar and Lang, Albert Georg and Buchner, Axel},
journal = {Behavior Research Methods},
issn = {1554351X},
number = {2},
pages = {175-191},
pmid = {17695343},
title = {{G*Power 3: A flexible statistical power analysis program for the social, behavioral, and biomedical sciences}},
volume = {39},
year = {2007}
}

@article{Chapman1994,
   author = {Loren J. Chapman and Jean P. Chapman and Timothy E. Curran and Michael B. Miller},
   issn = {02732297},
   issue = {2},
   journal = {Developmental Review},
   pages = {159-185},
   title = {Do Children and the elderly show heightened semantic priming? {H}ow to answer the question},
   volume = {14},
   year = {1994},
}

\end{document}